\documentclass[preprint,12pt]{article}
\usepackage{arxiv}

\usepackage{amssymb}
\usepackage[utf8]{inputenc}
\usepackage{amssymb,amsmath,array,amsthm}
\usepackage{mathtools}
\usepackage{amsmath}
\usepackage{amsfonts}
\usepackage{nicefrac}  
\usepackage{subfigure}
\usepackage{color}
\usepackage{xcolor}
\usepackage{mathtools}
\usepackage{graphicx}
\usepackage{colortbl}
\usepackage{booktabs} 
\usepackage{multirow,siunitx,etoolbox}
\usepackage{hyperref}
\newcommand{\rpm}{\sbox0{$1$}\sbox2{$\scriptstyle\pm$}
  \raise\dimexpr(\ht0-\ht2)/2\relax\box2 }

\title{Simple Graph Convolutional Networks}
\begin{document}
\author{
Luca Pasa$^{1,2}$, Nicolò Navarin$^{1,2}$, Wolfgang Erb$^{1}$, Alessandro Sperduti$^{1,2}$ \\
$^{1}$University of Padua - Department of Mathematics ``Tullio Levi-Civita''\\
via Trieste 63, 35121 Padua Italy\\
$^{2}$University of Padua - Human Inspired Technology Research Centre\\
Via Luzzatti 4, 35121 Padua Italy

}

\maketitle
\begin{abstract}
Many neural networks for graphs are based on the graph convolution operator,
proposed more than a decade ago.
Since then, many alternative definitions have been proposed, that tend to add complexity (and non-linearity) to the model.
In this paper, we follow the opposite direction by proposing simple graph convolution operators, that can be implemented in single-layer graph convolutional networks. 
We show that our convolution operators are more theoretically grounded than many proposals in literature, and exhibit state-of-the-art predictive performance on the considered benchmark datasets.

\end{abstract}


\section{Introduction}

In the last few years, there has been an increasing interest in machine learning models able to deal with graph-structured data, including kernel methods~\cite{Navarin2017} and neural networks~\cite{Scarselli2009}.
The idea of Graph Neural Networks (GNNs) is to define a neural architecture that follows the topology of the graph. Then a transformation is performed from the neurons corresponding to a vertex and its neighborhood to a  hidden representation, that is associated with the same vertex in another layer of the network. A new transformation is then performed for each hidden layer of the GNN. Each of these transformations depends on some parameters, that may be shared among all the vertices, obtaining Graph Convolutional Networks (GCNs).

All these models share the intuition that non-linearities are essential to obtain methods with high accuracy.
Recently, \cite{Wu2019} does put this concept into discussion, showing that removing the non-linearities from a popular GCN model actually did not impact much on the resulting predictive performance. 

In the conference paper \cite{Erb2020}, we took a further step in this direction. Starting from the theoretical foundations of graph convolution (GC), i.e. graph spectral filters, we defined a theoretically grounded  graph convolution layer.
We then built a single-layer graph neural network exploiting this layer, and showed
that the resulting Linear GCN  performed better than many approaches in literature on different benchmark semi-supervised node classification tasks.

In this paper, we extend the work in \cite{Erb2020}.
From the theoretical point of view, we revisit the definition of parametrized graph spectral filter and define three increasingly expressive graph convolutions (two of which are contributions of this paper). 
Moreover, we provide Rademacher complexity upper bounds for two of them.
From the experimental point of view, we include more datasets in our experimental comparison, and show that the proposed models achieve state-of-the-art predictive performance.
Finally, we include a detailed comparison on the computational requirements of our proposed methods, showing that two of our proposals are among the fastest methods in literature.

\section{Background}\label{sec:graphconvolution}
In the following, we denote scalars with lowercase letters, e.g. $x$, vectors with bold lowercase letters, e.g. $\mathbf{a}$, and matrices with bold uppercase letters, e.g. $\mathbf{M}$. When referring to the elements of a matrix, we use the row and column indices as subscripts, and the same letter used for the matrix in lowercase, i.e. $m_{ij}$ denotes the element at the $i$-th row, $j$-th column of $\mathbf{M}$. Further, we denote sets with uppercase letters, e.g. $S$.

Let $G=(V,E,\mathbf{X})$ be a graph, where $V=\{v_0, \ldots, v_{n-1}\}$ denotes the set of vertices (or nodes) of the graph, $E \subseteq V \times V$ is the set of edges, and $\mathbf{X} \in \mathbb{R}^{n\times {c}}$ is a multivariate signal on the graph nodes with the $i$-th row representing the attributes of $v_i$. We define $\mathbf{A} \in \mathbb{R}^{n \times n}$ as the adjacency matrix of the graph, with elements $a_{ij}=1 \iff (v_i,v_j)\in E$ and $a_{ij}=0$ otherwise. 

In this paper, we deal with the problem of semi-supervised learning on the nodes of the graph $G$, and we focus on neural networks models.
The first works extending neural networks to inputs in the graph domain
\cite{SperdutiStarita,Micheli2009,Scarselli2009} are based on the idea of aggregating the representation of a node and its neighbors, either in a recursive or a feed-forward (convolutive) way.
This idea has been re-branded later as \emph{graph convolution} or
\emph{neural message passing}.
In general, the main idea is to define the neural architecture following the topology of the graph. 
For each vertex a new hidden representation is computed thought an aggregation function that involves the vertex and its neighborhood. The aggregation function depends on some parameters, that may be shared among all the vertices.
Recently, many different graph convolutions have been proposed. In the following section, we discuss and derive in detail one of the most commonly adopted, while we review other graph convolutions in Section~\ref{sec:related}.

\subsection{Graph convolutions}
The derivation of the graph convolution operator originates from graph spectral filtering~\cite{Defferrard2016a,Hammond2009}.
Let us fix a graph $G$. Let $\mathbf{x}: V \rightarrow \mathbb{R}$ be a signal on the nodes $V$ of the graph $G$, i.e. a function that associates a real value to each node of $V$.
Since the number of nodes in $G$ is fixed (i.e. $n$) and the set $V$ is ordered, we can naturally represent every signal as a vector $\mathbf{x} \in \mathbb{R}^n$. In order to set up a convolutional network on $G$, we need the notion of a convolution $*_G$ between a signal $\mathbf{x}$ and a filter signal $\mathbf{f}$. However, as we don't have an inherent description of translation on $G$, it is not so obvious how to define the convolution directly in the graph domain. This operation is therefore usually defined in the spectral domain of the graph, using an analogy to classical Fourier analysis in which the convolution of two signals is calculated as the pointwise product of their Fourier transforms.

For this reason, we first provide a definition of the graph Fourier transform~\cite{Shuman2013}.
Let $\mathbf{L}$ be the (normalized) graph Laplacian, defined as:
\begin{equation}
    \mathbf{L}= \mathbf{I_n} - \mathbf{D}^{-\frac{1}{2}}\mathbf{A}\mathbf{D}^{-\frac{1}{2}}
\end{equation}
where $\mathbf{I_n}$ is the $n \times n$ identity matrix, and $\mathbf{D}$ is the degree matrix with entries given as
\begin{equation}
    d_{ij}= \left.
  \begin{cases}
    \sum_{k=0}^{n -1} a_{ik}, & \text{if } i=j \\
    0, & \text{otherwise}
  \end{cases}.
  \right.
\end{equation}
Since $\mathbf{L}$ is real, symmetric and positive semi-definite, we can compute its eigendecomposition as:
\begin{equation}
    \mathbf{L}=\mathbf{U}\mathbf{\Lambda}\mathbf{U^\top}
\end{equation}
where $\mathbf{\Lambda}= \text{diag}(\lambda_0,\ldots,\lambda_{n-1})$ is a diagonal matrix with the ordered eigenvalues of $\mathbf{L}$ as diagonal entries, and the orthonormal matrix $\mathbf{U}$ contains the corresponding eigenvectors $\{\mathbf{u}_0, \ldots, \mathbf{u}_{n-1}\}$ of $\mathbf{L}$ as columns.
In many classical settings of Fourier analysis, as for instance the Euclidean space or the torus, the Fourier transform can be defined in terms of the eigenvalues and eigenvectors of the Laplace operator.
In analogy, we consider now the eigenvectors $\{\mathbf{u}_0, \ldots, \mathbf{u}_{n-1}\}$ as the Fourier basis on the graph $G$ and the eigenvalues $\{\lambda_0,\ldots,\lambda_{n-1}\}$ as the corresponding graph frequencies. In particular, going back to our 
spatial signal $\mathbf{x}$, we can define its graph Fourier transform as:
\begin{equation}
\hat{\mathbf{x}}=   \mathbf{U^\top}\mathbf{x},
\end{equation}
and its inverse graph Fourier transform as:
\begin{equation}
\mathbf{x}=   \mathbf{U}\hat{\mathbf{x}}. 
\end{equation}
The entries $\hat{x}_i = \mathbf{x} \cdot \mathbf{u}_i$ are the frequency components or coefficients
of the signal $\mathbf{x}$ with respect to the basis function $\mathbf{u}_i$ and associated with the graph frequency $\lambda_i$. For this reason, $\hat{\mathbf{x}}$ can also be regarded as a distribution on the spectral domain of the graph, i.e. to each basis function $\mathbf{u}_i$ with frequency $\lambda_i$ a corresponding coefficient $\hat{x}_i$ is associated. 

\noindent Using the graph Fourier transform to switch between spatial and spectral domain, we are now ready to define the graph convolution between a filter $\mathbf{f}$ and a signal $\mathbf{x}$ as:
\begin{equation}
    \mathbf{f} *_G \mathbf{x} = \mathbf{U} \left ( \hat{\mathbf{f}}   \odot  \hat{\mathbf{x}} \right ) = \mathbf{U} \left ( \left ( \mathbf{U^\top}\mathbf{f} \right )  \odot \left ( \mathbf{U^\top} \mathbf{x} \right )\right ) \label{eq:spectralfilter1},
\end{equation}
where $\hat{\mathbf{f}}   \odot  \hat{\mathbf{x}} = (\hat{f}_0 \hat{x}_0, \ldots, \hat{f}_{n-1} \hat{x}_{n-1})$
denotes the component-wise Hadamard product of the two vectors $\hat{\mathbf{x}}$ and $\hat{\mathbf{f}}$. 

\noindent For graph convolutional networks, it is easier to design the filters $\mathbf{f}$ in the spectral domain as a distribution $\hat{\mathbf{f}}$, and then to define the filter $\mathbf{f}$ on the graph as 
$\mathbf{f}= \mathbf{U} \hat{\mathbf{f}}$. According to eq.~\eqref{eq:spectralfilter1}, for a given $\hat{\mathbf{f}}$ the application of the convolutional filter $\mathbf{f}$ to a signal $\mathbf{x}$ is given as:
\begin{equation} \label{eq:spectralfilter2}
     \mathbf{f} *_G \mathbf{x}= \mathbf{U} \left ( \left ( \mathbf{U^\top}\mathbf{U}\hat{\mathbf{f}} \right )  \odot \left ( \mathbf{U^\top}\mathbf{x} \right )\right )=
     \mathbf{U} \left (  \hat{\mathbf{f}}   \odot \left ( \mathbf{U^\top}\mathbf{x} \right )\right ).
\end{equation}
The Hadamard product $\hat{\mathbf{f}} \odot \hat{\mathbf{x}}$ can be formulated in matrix-vector notation as 
$\hat{\mathbf{f}} \odot \hat{\mathbf{x}} = \mathbf{\hat{F}} \hat{\mathbf{x}}$ by applying the diagonal matrix 
$\mathbf{\hat{F}} = \mathrm{diag}(\hat{\mathbf{f}})$, given by
$$(\mathbf{\hat{F}})_{ij} = 
\big(\mathrm{diag}(\hat{\mathbf{f}})\big)_{ij}=\begin{cases}
\hat{f}_i & \text{if }i=j\\
0 & \text{otherwise}
\end{cases},
$$
to the vector $\hat{\mathbf{x}}$. According to eq.~\eqref{eq:spectralfilter2}, we therefore obtain:
\begin{equation}
    \mathbf{f} *_G \mathbf{x}=\mathbf{U}\mathbf{\hat{F}}\mathbf{U^\top}\mathbf{x}.
    \label{eq:graphconv2}
\end{equation}

\noindent We can design the diagonal matrix $\mathbf{\hat{F}}$ and, thus, the spectral filter $\mathbf{f}$ in various ways. The simplest way would be to define 
$\mathbf{f}_{\boldsymbol{\theta}}$ as a parametric filter, i.e. use $\mathbf{\hat{F}}_{\boldsymbol{\theta}} = \mathrm{diag}(\boldsymbol{\theta})$, where $\boldsymbol{\theta}=(\theta_0, \ldots, \theta_{n-1})^\top$ is a completely free vector of filter parameters that can be learned by the neural network. However, such a filter grows in size with the data, and it is not well suited for learning.

\noindent A better option pursued in this work is to use a polynomial parametrization based on powers of the spectral matrix $\mathbf{\Lambda}$ for the filter, such as:
\begin{equation}
    \mathbf{\hat{F}}_{\boldsymbol{\theta}} = \sum_{i=0}^{k} \theta_i \mathbf{\Lambda}^i.
    \label{eq:powerserielambda}
\end{equation}
This filter has $k+1$ parameters $\{\theta_0, \ldots, \theta_k \}$ 
to learn, and it is spatially $k$-localized on the graph. One of the main advantages 
of this filter is that we can formulate it explicitly in the graph domain. Recalling
the eigendecomposition $\mathbf{L}=\mathbf{U}\mathbf{\Lambda}\mathbf{U}^\top$
of the graph Laplacian, eq.~\eqref{eq:graphconv2} and eq.~\eqref{eq:powerserielambda} combined give:

\begin{align}
        \mathbf{f}_{\boldsymbol{\theta}} *_G \mathbf{x} &=  \notag
        \mathbf{U} \mathbf{\hat{F}}_{\boldsymbol{\theta}} \mathbf{U^\top}\mathbf{x} 
        =  \sum_{i=0}^k \theta_i  \mathbf{U} \mathbf{\Lambda}^i \mathbf{U^\top} \mathbf{x}  \\ 
        &= \sum_{i=0}^k \theta_i (\mathbf{U} \mathbf{\Lambda} \mathbf{U^\top})^i \mathbf{x} = \sum_{i=0}^k \theta_i \mathbf{L}^i \mathbf{x}.
    \label{eq:graphconv3}
\end{align}

\noindent Note that the computation of the eigendecomposition of the graph Laplacian $\mathbf{L}$ (the cost is of order $O(n^3)$) is feasible only for relatively small graphs (with some thousands nodes at most).
Real-world problems involve however graphs with hundreds of thousands or even millions nodes: in these cases, the computation of the eigendecomposition of $\mathbf{L}$ is prohibitive and a filter of
the form of eq.~\eqref{eq:graphconv3} has clear advantages compared to a spectral filter given in the 
form of eq.~\eqref{eq:graphconv2}.

The parametrization of the polynomial filter eq.~\eqref{eq:powerserielambda} is given 
in the monomial basis. Alternatively, \cite{Defferrard2016a} proposes to use Chebyshev polynomials as a polynomial basis. In general, the usage of a Chebyshev basis improves the stability in numerical approximation.

In \cite{Kipf2016a}, the authors propose to fix the order $k=1$ in eq.~\eqref{eq:powerserielambda} to obtain a linear first order filter for each graph convolutional layer in a neural network. These simple convolutions can then be stacked in order to improve the discriminatory power of the resulting network.
The resulting convolution operator in \cite{Kipf2016a} is defined as:
\begin{equation}
    \mathbf{f}_{\theta} *_G \mathbf{x} = \theta(\mathbf{I_n} +\mathbf{D}^{-\frac{1}{2}} \mathbf{A}\mathbf{D}^{-\frac{1}{2}})\mathbf{x} = \theta(2 \mathbf{I_n} - \mathbf{L})\mathbf{x}.
\end{equation}
The authors in \cite{Kipf2016a} additionally use a renormalization trick to
limit the eigenvalues of the resulting matrix: they replace $\mathbf{I_n}+ \mathbf{D}^{-\frac{1}{2}} \mathbf{A}\mathbf{D}^{-\frac{1}{2}}$ by $\tilde{ \mathbf{D}}^{-\frac{1}{2}} \tilde{\mathbf{A}}\tilde{\mathbf{D}}^{-\frac{1}{2}}$, where $\tilde{\mathbf{A}}=\mathbf{A}+ \mathbf{I_n}$ and $(\mathbf{D})_{ii}=\sum_{j=0}^n (\tilde{\mathbf{A}})_{ij}$. In this way, the spectral filter $\mathbf{f}_{\boldsymbol{\theta}}$ is not build upon the spectral decomposition of the graph Laplacian $\mathbf{L}$ but on the eigendecomposition of the perturbed operator $\tilde{ \mathbf{D}}^{-\frac{1}{2}} \tilde{\mathbf{A}}\tilde{\mathbf{D}}^{-\frac{1}{2}}$. 

\noindent Applying this convolution operator to a multivariate signal $\mathbf{X} \in \mathbb{R}^{n \times c}$ and using $m$ filters, we obtain the following definition for a single graph convolutional layer:
\begin{equation} \label{eq:singleGCN}
    \mathbf{H}= \tilde{ \mathbf{D}}^{-\frac{1}{2}} \tilde{\mathbf{A}}\tilde{\mathbf{D}}^{-\frac{1}{2}} \mathbf{X} \mathbf{\Theta},
\end{equation}
where $\mathbf{\Theta} \in \mathbb{R}^{c \times m}$.
This convolutional operation has complexity $O(|E|mc)$.
To obtain a Graph Convolutional Network (GCN), several graph convolutional layers are stacked and interleaved by a nonlinear activation function, typically a ReLU.\\
If $\mathbf{H}^{(0)}=\mathbf{X}$, then, based on the single layer convolution in eq.~\eqref{eq:singleGCN}, we obtain the following recursive definition for the $k$-th graph convolutional layer:
\begin{equation}
    \mathbf{H}^{(k)}= ReLU(\tilde{ \mathbf{D}}^{-\frac{1}{2}} \tilde{\mathbf{A}}\tilde{\mathbf{D}}^{-\frac{1}{2}} \mathbf{H}^{(k-1)} \mathbf{\Theta}).
    \label{eq:gcn}
\end{equation}
Although GC can be applied also to other settings, we will from now on focus on the task of multiclass classification. In the last GC layer (say the $l$-th) the ReLU activator is replaced by a softmax classifier (that is a multinomial logistic regression) to output the predictions:
\begin{equation}
    \mathbf{Y}= softmax(\tilde{ \mathbf{D}}^{-\frac{1}{2}} \tilde{\mathbf{A}}\tilde{\mathbf{D}}^{-\frac{1}{2}} \mathbf{H}^{(l-1)} \mathbf{\Theta}).
\end{equation}
\subsection{Simple Graph Convolution}
In \cite{Wu2019}, a simplification of the convolution operator in eq.~\eqref{eq:gcn} is proposed, dubbed Simple Graph Convolution (SGC). The idea is that perhaps the nonlinear operator introduced by GCNs is not essential. However, stacking multiple GC layers has an important effect on the locality of the learned filters, i.e. after $k$ GC layers, the hidden representation of a vertex considers information coming from the vertices up to distance $k$, i.e. the filters on the $k$-th layer are $k$-localized.
Let us rewrite, for ease of notation:
\begin{equation}
\mathbf{S}=\tilde{\mathbf{D}}^{-\frac{1}{2}} \tilde{\mathbf{A}}\tilde{\mathbf{D}}^{-\frac{1}{2}},
\label{eq:S}
\end{equation}
then a GC layer as defined in eq.~\eqref{eq:gcn} (not considering the ReLU non-linearity) becomes $\mathbf{H}^{(i)}= \mathbf{S} \mathbf{H}^{(i-1)}\mathbf{\Theta}$.
If we stack $k$ such layers with no non-linearity, and we apply a \emph{softmax} classifier at the end, the output after $k$ hidden layers is:
\begin{equation}
    \mathbf{Y}=softmax(\mathrlap{\overbrace{\phantom{\mathbf{S}\ldots \mathbf{S}}}^{k}}\mathbf{S}\ldots \mathbf{S}\mathbf{X}\mathbf{\Theta}^{(0)} \ldots \mathbf{\Theta}^{(k-1)}).
\end{equation}
Since the SGC model is linear, we can reparametrize it as $\mathbf{\Theta}=\mathbf{\Theta^{(0)}} \ldots \mathbf{\Theta^{(k-1)}}$ obtaining:
\begin{equation}
    \mathbf{Y}=softmax(\mathbf{S}^k\mathbf{X}\mathbf{\Theta}).
    \label{eq:sgc}
\end{equation}
The great advantage of this model is a reduced number of parameters compared to classical graph convolution. Moreover, $\mathbf{S}^k$ can be computed only once, with a drammatic speedup compared to GCNs.
\subsubsection*{Interpretation as Logistic Regression over graphs}
The SGC formulation has an interesting interpretation. We can think about having a fixed feature extractor/representation for the graph \mbox{($\bar{\mathbf{X}}=\mathbf{S}^k\mathbf{X}$)} and a simple multinomial logistic regression applied to it: $$\mathbf{Y}=softmax(\bar{\mathbf{X}}\mathbf{\Theta}).$$ 
The training of the model reduces to training a standard softmax classifier.

\section{Proposed Graph Convolutions}
In this section, we present the main contributions of this paper. We first introduce (an approximate) exponential graph filter. Then, we present the Linear Graph Convolution (LGC), that was proposed in the preliminary conference version of this work~\cite{Erb2020}, and that is based on a combination of monomial kernels and that is more expressive than EGC.
We present Rademacher complexity upper bounds for these two convolutions. Since we propose to use a single graph convolution layer, these bounds can be directly used to estimate the generalization error.
Finally, we define an even more expressive convolution, the Hyper-LGC, that depends on a number of parameters that is linear in the size of the graph. 
This is implemented
via a parametrized (and learnable) function that generates the values of the coefficents depending on the considered input vertex.

\subsection{Exponential Graph Convolution}

In this section, we introduce the Exponential Graph Convolution (EGC) operator, based on the coefficients of the exponential power series. 
We revisit the definition of parametrized filter in eq.~\eqref{eq:powerserielambda}, considering an exponential filter instead of a polynomial one:

\begin{equation}
    \mathbf{\hat{F}}_{\beta} = e^{\beta\boldsymbol{\Lambda}}= \sum_{i=0}^\infty \frac{\beta^i}{i!} \boldsymbol{\Lambda}^i.
    \label{eq:powerserielambdaexponential}
\end{equation}
With this filter the convolution in the graph domain is given as follows:

\begin{equation*}
     \mathbf{f}_{\beta} *_G \mathbf{x} = e^{\beta\mathbf{L}}\mathbf{x} = \sum_{i=0}^\infty \frac{\beta^i}{i!} \mathbf{L} \mathbf{x}.
\end{equation*}

Truncating the series to a maximum number $k$ and applying this filter to a multivariate signal and $m$ outputs, we obtain one layer of the Exponential Graph Convolution (EGC) as
\begin{equation}
\mathbf{H}^{(k)} = \sum_{i=0}^k \frac{\beta^i}{i!} \mathbf{L}^i\mathbf{X}\mathbf{\Theta},
\label{eq:EGC_H}
\end{equation}
where $\beta$ and $\mathbf{\Theta}$ are the parameters to be learned.

\noindent Note that in the limit $k \to \infty$, we get
 \begin{equation}
 \mathbf{H}^{(\infty)} =\lim_{k \to \infty} \mathbf{H}^{(k)} = e^{\beta \mathbf{L}}\mathbf{X}\mathbf{\Theta}.
     \label{eq:EGC_H_INFTY}
 \end{equation}

\noindent This can be derived from the following approximation error of our truncated EGC with respect to $\mathbf{H}_{\infty}$: 
\begin{equation}
\| \mathbf{H}^{(\infty)} - \mathbf{H}^{(k)} \| = \left\| \sum_{i = k+1}^\infty \frac{\beta^i}{i!} \mathbf{L}^i\mathbf{X}\mathbf{\Theta}\right\| \leq \frac{|\beta|^{k+1}\|\mathbf{L}\|^{k+1}}{(k+1)!} \frac{\|\mathbf{X}\mathbf{\Theta}\|}{1- |\beta|\|\mathbf{L}\|/(k+2)} ,
\label{eq:EGC_H_DIFF}
\end{equation}
where $\|\cdot\|$ denotes the spectral norm for matrices. In particular, this estimate guarantees that for a truncation number $k$ large enough the polynomial EGC model mimics a graph convolution with an exponential kernel. As the solution of the diffusion equation on the graph is determined by the exponential kernel $e^{\beta \mathbf{L}}$, the input and the output layer in the EGC model are linked by a diffusion process on the graph nodes. Moreover, during the training process the diffusion rate $\beta$ is optimally adjusted to the given training set. This yields a first improvement over the SGC model introduced above in which the propagation matrix $\mathbf{S}^k$ was a priori fixed and just the weight matrix $\mathbf{\Theta}$ is determined during the learning process.

{Instantiating the single-layer neural network with the proposed EGC filter, we obtain  as final model:}
\begin{equation}
            \mathbf{Y}=softmax(\sum_{i=0}^k \frac{\beta^i}{i!} \mathbf{L}^i\mathbf{X}\mathbf{\Theta}).
            \label{eq:EGC}
\end{equation}

\subsection{Linear Graph Convolution}
By considering the EGC formulation in eq.~\eqref{eq:EGC},
we observe that the single parameter $\beta$ determines the weights assigned to all the components in the summation (via the exponential series expansion).  In this way, possible graph convolutions in the network are limited to those described by exponential filters. In particular, these filters act as low pass filters that emphasize the contributions of the low order monomials in the representation of $\mathbf{H}$ more than the high order ones. This might be a restriction for some applications. 

To allow larger families of polynomial filters in a network layer and to increase the expressive power of the convolution operator, we therefore propose in this section Linear Graph Convolutions (LGCs). The main idea is to replace the terms $\frac{\beta^i}{i!}$ with learnable parameters, one for each $i$, obtaining the following formulation for a single layer:
\begin{equation}
         \mathbf{H}=\sum_{i=0}^k \alpha_i \mathbf{L}^i\mathbf{X}\mathbf{\Theta}.
            \label{eq:LGC_H}
\end{equation}

\noindent Similarly as before, we can define a single-layer LGC neural network as:
\begin{equation}
            \mathbf{Y}=softmax(\sum_{i=0}^k \alpha_i  \mathbf{L}^i\mathbf{X}\mathbf{\Theta}).
            \label{eq:LGC}
\end{equation}

Note that in this work we derived the EGC and LGC formulation in terms of spectral filters based on the graph Laplacian $\mathbf{L}$, while in the conference paper~\cite{Erb2020} we proposed the same formulation as an extension of the SGC operator. In fact, by using the perturbed operator $\mathbf{S}$ instead of $\mathbf{L}$ and fixing the coefficients $\alpha_0 = \ldots = \alpha_{k-1} = 0$, and $\alpha_k = 1$, we see that the SGC operator fits as well in this more general LGC framework, if we ignore the slightly different normalization for $\mathbf{S}$. Compared to the fixed SGC scheme, and the EGC scheme with one additional parameter $\beta$, the more flexible LGC formulation allows to learn $k+1$ coefficients for the convolution in the network layer. 

This added expressiveness does not allow us to bound the approximation error introduced with respect to the version with $k=\infty$. In fact, having no constraints on the $\alpha_i$ parameters, each term of the summation can potentially significantly contribute to the final representation.

{ \subsection{Rademacher complexity for LGC and EGC}
The simple single-layered structure of the EGC and LGC networks allows to obtain explicit estimates for their Rademacher complexity, a measure for the learnability of function classes in the respective networks. 

For a set $\mathcal{F}$ of real-valued signals $f: V \to \mathbb{R}$ on the graph and a sampling set $U = \{u_1, \ldots, u_L\} \subset V$ (usually i.i.d. random nodes distributed according to a given probability measure on the domain $V$) the empirical Rademacher complexity of $\mathcal{F}$ w.r.t. $U$ is defined as
\[\hat{\mathcal{R}}(\mathcal{F}) := \mathbb{E}_{\epsilon} \left[\sup_{f \in \mathcal{F}} \frac{1}{L} \sum_{\ell = 1}^L \epsilon_{\ell} f(u_{\ell})\right],\]
where $\mathbb{E}_{\epsilon}$ denotes the expectation w.r.t. a uniform distribution of $\epsilon \in \{-1,1\}^L$. \\

{\noindent\textbf{Theorem 1.}} Let $a,b>0$, $ \sup_{j,j'} |\mathbf{X}_{j,j'}| \leq M$, and consider the set
\[\mathcal{F}_{\mathrm{LGC}} = \left\{\mathbf{Y}=\sigma(\sum_{i=0}^k \alpha_i  \mathbf{L}^i \mathbf{X}\mathbf{\Theta}) \; | \; \|\mathbf{\alpha}\|_{\infty}\leq a,\; \|\mathbf{\Theta}\|_{1} \leq b\right\},\]
where $\sigma$ is any Lipschitz-continuous activation function with Lipschitz constant $\Lambda$ and $\Theta \in \mathbb{R}^c$. Then the Rademacher complexity of $\mathcal{F}_{\mathrm{LGC}}$ w.r.t. any sampling set $U \subset V$ of size $L$ is bounded by
\begin{equation}\label{eq:RademacherLGC}
\hat{\mathcal{R}}(\mathcal{F}_{\mathrm{LGC}})
\leq \frac{b M \Lambda}{\sqrt{L}} \left(\sum_{i=0}^k a \|\mathbf{L}\|_1^i\right).
\end{equation}

{\noindent \textit{Proof of Theorem 1.}}
Without loss of generality we can assume that $u_1 = v_1,$ $ \ldots, u_L = v_L$. Since the activation function $\sigma$ is Lipschitz with constant $\Lambda > 0$, we can use the contraction property of the Rademacher complexity and obtain
\begin{align*}
\hat{\mathcal{R}}(\mathcal{F}_{\mathrm{LGC}}) &\leq \frac{\Lambda}{L} \mathbb{E}_{\epsilon}\left[\sup_{ \|\mathbf{\alpha}\|_{\infty}\leq a} \sup_{\|\mathbf{\Theta}\|_{1} \leq b} \sum_{\ell = 1}^L \epsilon_{\ell}\left(\sum_{i=0}^k \alpha_i  \mathbf{L}^i \mathbf{X}\mathbf{\Theta} \right)_{\ell}\right]\\ & \leq 
\sup_{ \|\mathbf{\alpha}\|_{\infty}\leq a} \sup_{\|\mathbf{\Theta}\|_{1} \leq b} \frac{\Lambda}{L} \left\|\left(\sum_{i=0}^k \alpha_i  \mathbf{L}^i \right) \mathbf{X} \mathbf{\Theta} \right\|_{\infty} \mathbb{E}_{\epsilon} \left| \sum_{\ell = 1}^L \epsilon_{\ell} \right| \\
&\leq \frac{b M \Lambda }{L} \left(\sum_{i=0}^k a \|\mathbf{L}\|_1^i\right) \sqrt{L}.
\end{align*} 
Here, in the last step we used Jensen's inequality. \qed \\

As expected, this theoretical estimate indicates that the bound on the Rademacher complexity of the LGC increases as soon as the degree $k$ of the polynomial kernel increases. Similarly as in Theorem 1, we can further show the following estimate for the Rademacher complexity of the EGC network.\\

{\noindent\textbf{Theorem 2.}} Let $a,b>0$, $ \sup_{j,j'} |\mathbf{X}_{j,j'}| \leq M$, and consider the set
\[\mathcal{F}_{\mathrm{EGC}} = \left\{\mathbf{Y}=\sigma(\sum_{i=0}^k \frac{\beta^i}{i!}  \mathbf{L}^i \mathbf{X}\mathbf{\Theta}) \; | \; |\beta|\leq a,\; \|\mathbf{\Theta}\|_{1} \leq b\right\},\]
where $\sigma$ is any Lipschitz-continuous activation function with Lipschitz constant $\Lambda$ and $\Theta \in \mathbb{R}^c$. Then the Rademacher complexity of $\mathcal{F}_{\mathrm{EGC}}$ w.r.t. any sampling set $U \subset V$ of size $L$ is bounded by
\begin{equation}\label{eq:RademacherEGC}
\hat{\mathcal{R}}(\mathcal{F}_{\mathrm{EGC}})
\leq \frac{b M \Lambda}{\sqrt{L}} 
e^{a \|\mathbf{L}\|_1}.
\end{equation}

Notice that the bound for EGC in eq.~\eqref{eq:RademacherEGC} is a better bound compared to the one for LGC in eq.~\eqref{eq:RademacherLGC} since it exploits the property of the coeffiecients of EGC being defined as a power series.

What is more important, however, is that the above bounds can be directly applied to the proposed models due to their simple one-layer structure. This is not the case for more complex architectures where two or more layers are stacked. For these type of networks, it is difficult to derive meaningful bounds since a bound on a single layer should be combinatorially reused 
leading to too large upper bounds.}

\subsection{Hyper-LGC}

EGC and LGC convolutions, presented in the previous sections, assign a single weight to each term in the summation.
LGC aimed at increasing the expressiveness of EGC exploiting multiple (i.e., $k+1$) weighting parameters compared to the single parameter $\beta$ of EGC, c.f. eq.~\eqref{eq:EGC} and eq.~\eqref{eq:LGC}.
To increase the expressiveness of the defined convolution even further, it would be possible to define a convolution with one parameter for each node and each layer. This convolution would be defined as:
\begin{equation}
\mathbf{H}=\sum_{i=0}^k \boldsymbol{\alpha}_i \odot \mathbf{L}^i\mathbf{X}\mathbf{\Theta},
\end{equation}
where $\boldsymbol{\alpha}_i \in \mathbb{R}^n$ is a vector of weights, one for each node and for each $i$.
Such definition, while allowing each node to aggregate information coming from multiple terms of the sum in a different way, would probably result in overfitting due to the high number of parameters and the lack of any regularization mechanism such as weight sharing.\\
We can, however, start from this intuition and define a mechanism to reduce the number of parameters. 
We propose to define the weight vectors $\boldsymbol{\alpha}_i$ as the output of a function implemented by a neural network. The number of parameters of such network will be lower with respect to the number of nodes, thus forcing the exploitation of locality and feature similarity in the graph domain.
We propose to parametrize such function on the input features propagated  via the diffusion operator at each $i$-hop step. Thus, the resulting function is a graph convolutional neural network itself. Moreover, we implement it using a gating mechanism that, for each node, modifies the \emph{base} parameter $\alpha_i$.

This idea is inspired by the Hypernetworks \cite{ha2016hypernetworks}. Hypernetworks were introduced in the context of an RNN and CNN that were used to generate the weights of a  primary model that computes the actual task. However, the idea of having one network to predict the weights of another was proposed earlier and has reemerged multiple times~\cite{klein2015dynamic, riegler2015conditioned, jia2016dynamic}.\\
The Hyper-LGC (hLGC) is defined as follows:

\begin{equation}
    f_i(\mathbf{L}^i\mathbf{X}) = \sigma({ReLu}(\mathbf{L}^i\mathbf{X} \cdot \mathbf{W}_1^{(i)})\cdot \mathbf{W}_2^{(i)})\alpha_i, \;\; \mathbf{W}_1^{(i)}\in \mathbf{R}^{s \times \frac{s}{2}}, \mathbf{W}_2^{(i)}\in \mathbf{R}^{\frac{s}{2}\times 1},\nonumber
\end{equation}
\begin{equation}
            \mathbf{Y}=softmax(\sum_{i=0}^k (\mathbf{L}^i\mathbf{X}\mathbf{\Theta} \odot f_i(\mathbf{L}^i\mathbf{X}))).
            \label{eq:hLGC}
\end{equation}

where $f()$ is a  neural network that returns a multiplicative factor value for each node. In the following, we refer to $f()$ as  the hyper model.
Note that, differently from the common Hyper Neural Network previously proposed in literature, the proposed architecture uses a simpler hyper model. Indeed, the adopted one is a simple network, while in the hyper neural network framework, it is common to use as hyper network a model that has a similar structure than the primary model (e.g., the hyper LSTM and the Hyper CNN proposed in \cite{ha2016hypernetworks}).

An interesting feature of the hyper networks is that their particular structure allows to overtake the limitation imposed by the weight sharing mechanism. 
Our approach slightly differs from the typical hyper-net mechanism. In fact the hLGC uses the hyper model $f()$ just to relax the weight sharing limitation.Indeed, hLGC does not delegate the management of all weights to the hyper network $f()$, while it maintains most of its weights shared among all graph nodes. The model exploits the hyper model just to create a multiplicative factor that allows the re-scaling  of the computed embedding.

\subsection{Differences with respect to SGC}
Comparing our proposed convolutions (EGC, LGC and hLGC) to SGC, and ignoring the difference in the adopted adjacency matrix function, i.e., $\mathbf{S}$ defined in eq.~\eqref{eq:S} and the graph Laplacian $\mathbf{L}$ (see Section \ref{sec:alternatives}), we see that for SGC the output depends just on the $k$-th term in the summation, i.e. the $k$-th exponentiation of the matrix $\mathbf{S}$ (see eq.~\ref{eq:sgc}).The terms with order lower than $k$ are not directly considered in the computed representation.
This is in contrast with our proposal that, based on approximation theory, we consider all powers of $\mathbf{S}$ up to a maximum degree $k$. The key difference of our proposed convolutions with respect to SGC is that in our proposals each power of $\mathbf{S}$ up to degree $k$ directly contributes to the output and the corresponding coefficients can be adjusted according to the given data.

\subsection{Alternative definition of graph convolution operators}
\label{sec:alternatives}
Up to now,
we presented different definitions of the graph convolutions based on the graph Laplacian~\cite{Sandryhaila2013}. However, in equations~\eqref{eq:EGC}, \eqref{eq:LGC} and \eqref{eq:hLGC},
we can also replace the graph Laplacian $\mathbf{L}$ with the re-normalized operator $\mathbf{S}$ in eq.~\eqref{eq:S}, as for instance done in the SGC model of eq.~\eqref{eq:sgc}. 
Moreover, if the considered graph is undirected, it is possible to define the graph Fourier transform in terms of its (normalized) adjacency matrix instead.
In fact, the adjacency matrix $\mathbf{A}$ of undirected graphs is real and symmetric, and thus its eigen-decomposition can always be computed.
Finally, it is also possible to consider the perturbed adjacency matrix $\mathbf{S}$ defined in eq.~\eqref{eq:S}.
The derivations of the graph convolutions remain similar as reported above.
Even though we see similar results, the formulation considering the normalized adjacency matrix seems to be more robust.

\subsection{Computational complexity}
SGC in eq.~\eqref{eq:sgc} is very efficient compared to other convolutions based on message passing, e.g. GCN in eq.~\eqref{eq:gcn}, because it is possible to precompute the term $\mathbf{S}^k\mathbf{X}$.
Considering our three proposed models in equations ~\eqref{eq:EGC}, \eqref{eq:LGC} and \eqref{eq:hLGC}, we can notice that for all of them the terms $\mathbf{L}^i\mathbf{X}$ can be precomputed as well. Thus, the computational requirements of our proposed convolutions are comparable to the ones of SGC.
While the asymptotic complexity of SGC, of our proposed methods, and other convolutions based on message passing is the same, in practice SGC as well as our proposed convolutions can be significantly faster compared to, for instance, the very popular GCN (see section~\ref{sec:times}).

\section{Related Works\label{sec:related}}
In the last few years several models inspired by the graph convolution idea have been proposed.
We already discussed some methods that are closer to our formulation in Section~\ref{sec:graphconvolution}. In this section, we detail other methods in literature that are relevant for historical reasons or that we use as comparison in our experiments.

Scarselli et al. \cite{Scarselli2009} proposed a  transition function on a graph vertex $v$ that at time $0\leq t$ is defined as:
\begin{equation}
\mathbf{h}^{t+1}_v =\sum_{u \in \mathcal{N}(v)} f(\mathbf{h}^{t}_u,\mathbf{x}_v,\mathbf{x}_u),
\label{eq:scarsellirec}
\end{equation}
where $f$ is a parametric function whose parameters have to be learned (e.g. a neural network) and are shared among all the vertices.
This transition function is part of a recurrent system. It is defined as a contraction mapping, thus the system is guaranteed to converge to a fixed point, i.e., a representation that does not depend on the particular initialization of the weight matrix $\mathbf{H}^0$.
The final representation for each vertex is computed from the last representation and the original vertex labels as follows:
\begin{equation}
\mathbf{o}^t_v=g(\mathbf{h}^t_v,\mathbf{x}_v),
\end{equation}
where $g()$ is another neural network.
The work \cite{Li2015b} modified the model proposed in \cite{Scarselli2009} by removing the constraint for the recurrent system to be a contraction mapping, and by replacing the recurrent units with GRUs.

\noindent Micheli~\cite{Micheli2009} proposed a model referred to as Neural Network for Graphs (NN4G). In the first layer, a transformation over vertex labels is computed:
\begin{equation}
\mathbf{H}^{(1)}=\sigma \left ( \mathbf{X} \mathbf{\bar{W}}^{(1)} \right ),
\label{eq:micheliconv1}
\end{equation}
where $\bar{\mathbf{W}}^{(1)}$ are the weights connecting the original labels $\mathbf{X}$ to the current layer, and $\sigma$ is a non-linear function applied element-wise.
The graph convolution is then defined for the $(i+1)$-th layer as: 
\begin{equation}
\mathbf{H}^{(i+1)}= \sigma \left ( \mathbf{X} \mathbf{\bar{W}}^{(i+1)} + \sum_{k=1}^{i}\mathbf{A} \mathbf{H}^{(k)} \mathbf{\hat{W}}^{(i+1,k)}\right ),
\label{eq:micheliconvmatrix}
\end{equation}
where ${i={0,\ldots,l-1}}$ (and $l$ is the number of layers), {$\mathbf{\bar{W}}^{(i+1)} \in \mathbb{R}^{d \times c_{i+1}}$}, {$\mathbf{\hat{W}}^{({i+1},k)} \in \mathbb{R}^{c_{k} \times c_{i+1}}$}, $\mathbf{H}^{(k)} \in \mathbb{R}^{n \times c_k}$, $c_i$ is the size of the hidden representation at the $i$-th layer.
\noindent The convolution in 
eq.~\eqref{eq:micheliconvmatrix}
is part of a multi-layer architecture, where each layer's connectivity resembles the topology of the graph, and the training is layer-wise.

Duvenaud et al.~\cite{Duvenaud2015b} proposed a hierarchical approach 
similar to NN4G and inspired by circular fingerprints in chemical structures. While NN4G \cite{Micheli2009} adopted Cascade-Correlation for training, Duvenaud et al.~\cite{Duvenaud2015b} proposed to use end-to-end back-propagation.
 ECC \cite{simonovsky2017dynamic} 
 is an improvement of the method proposed by Duvenaud et al., weighting the sum over the neighbors of a vertex by weights conditioned by the edge labels.

Zhang et al. \cite{Zhang2018} proposed a  propagation scheme for vertices' representations based on the random-walk graph Laplacian, similar to the one presented in eq.~\eqref{eq:gcn}. 
Authors state that the choice of normalization does not significantly affect the results.
The graph convolution proposed by Zhang et al. has been recently extended with an hyper-parameter controlling the neighborhood distance considered in the convolution operation \cite{DinhNavarin2018}.

PATCHY-SAN
\cite{niepert2016learning} follows a more straightforward approach to define convolutions on graphs, that is conceptually closer to convolutions defined over images. First, it selects a fixed number of vertices from each graph, exploiting a canonical ordering on graph vertices. Then, for each vertex, it defines a fixed-size neighborhood (of vertices possibly at distance greater than one), exploiting the same ordering. 
This approach requires to compute a canonical ordering over the vertices of each input graph, that is a problem as complex as the graph isomorphism (no polynomial-time algorithm is known). Authors resort to the tool Nauty~\cite{Mckay2014} that, while being practically fast compared to other solutions, introduces a computational bottleneck for the method.

Diffusion CNN 
\cite{atwood2016diffusion} defines a different graph convolution (i.e. diffusion-convolution) that incorporates in the definition of graph convolution the diffusion operator, i.e. the multiplication of the input representation with a power series of the degree-normalized transition matrix.


Graph Attention Networks (GAT)~\cite{Velickovic2017} exploit a different convolution operator based on masked self-attention. 
The idea is to replace the adjacency matrix in the convolution with a matrix of attention weights:
\begin{equation}
    \mathbf{H}^{(i+1)}=\sigma(\mathbf{B}^{(i+1)}\mathbf{H}^{(i)}\mathbf{\Theta}) \label{eq:gat1},
\end{equation}
where $0 \leq i < l$ (the number of layers), $\mathbf{H}^{(0)}=\mathbf{X}$, and the $u,v$-th element of $\mathbf{B}^{(i+1)}$ is defined if $(u,v)\in E$ as: 
\begin{equation}
    b_{u,v}^{(i+1)}=\frac{exp(LeakyRELU(\mathbf{w'}^\top [\mathbf{W} \mathbf{h}^{(i)}_u || \mathbf{W} \mathbf{h}^{(i)}_v]))}{
    \sum_{k \in \mathcal{N}(u)} exp(LeakyRELU(\mathbf{w'}^\top [\mathbf{W} \mathbf{h}^{(i)}_u || \mathbf{W} \mathbf{h}^{(i)}_k]))}, \label{eq:gat2}
\end{equation}
$0$ otherwise. The vector $\mathbf{w'}$ and the matrix $\mathbf{W}$ are learnable parameters.
Authors propose to use multi-head attention to stabilize the training. While it may be more complex to train, GAT allows to weight differently the neighbors of a node, thus it is a very expressive graph convolution.
Fast GCN~\cite{Chen2018} uses node sampling to define a fast convolution operator, suited for the inductive setting.
Graph Isomorphism Networks (GIN)~\cite{Xu2018HowPA} and \cite{DinhNavarin2018} adopt a more powerful graph convolution operator. LNet and AdaLNet ~\cite{Liao2019} exploit filters learned on an approximation of the Laplacian matrix. Deep Graph InfoMax (DGI)~\cite{Velickovic2019} trains a GCN in an unsupervised setting to obtain general node embeddings.
GNN with ARMA filters (ARMA) \cite{MariaBianchi2019} defines an ARMA filter for graph convolution.

\section{Results}
In this section, we compare the proposed graph convolutional layers against several state-of-the-art alternatives on five real-world node classification datasets.

\subsection{Dataset}
We empirically validated the proposed convolutions on four widely adopted datasets of node classification: Citeseer, Cora, Pubmed, and Reddit. Each dataset is a graph, and in the first three of them, nodes represent documents and node features are sparse bag-of-words feature vectors. Specifically, in Citeseer, Cora, and Pubmed the task requires to classify the research topics of papers. Each node represents a scientific publication described by a 0/1-valued word vector indicating the absence/presence of the corresponding word from a dictionary.
In Reddit dataset the task involves 
the classification of Reddit posts.
Each node is a post, and the node label is the community, or “subreddit”, that a post belongs to. The authors sampled 50 large communities and built a post-to-post graph, connecting posts if the same user comments on both. Relevant statistics about the datasets are reported in Table~\ref{tab:dataset}.
\setlength\tabcolsep{3pt}
\begin{table}
  \centering
  \begin{tabular}{l|c|c|c|c|c}
    \toprule
    {Dataset} &  {\#Classes} & {\#Edges} & {\#Train} & {\#Val} & {\#Test}   \\
    \midrule
    {\textbf{Citeseer}} & $6$ & $9228$ & $120$ & $500$ & $1000$ \\
    \hline
    {\textbf{Cora}}  & $7$ & $10556$ & $140$ & $500$ & $1000$ \\
    \hline
    {\textbf{Pubmed}}  & $3$ & $88651$ & $60$ & $500$ & $1000$ \\
    \hline
    {\textbf{Reddit}}  & $41$ & $114848857$ & $153431$ & $23831$ & $55703$ \\

    \bottomrule
  \end{tabular}
  \caption{Datasets statistics. The columns {\#Train}, {\#Val}, and {\#Test} report the number of nodes in the training, validation and test sets, respectively.}
  \label{tab:dataset}
\end{table}

\subsection{Experimental setting and implementation details}

We developed all the models involved in the comparison using Deep Graph Library (DGL)~\cite{wang2019dgl}. As baseline models, we considered the SGC (see eq.~\eqref{eq:sgc}), the GAT (see eqs.~\eqref{eq:gat1}-\eqref{eq:gat2}) and the GCN (see eq.~\eqref{eq:gcn}) convolutions. For these models we exploit the implementation provided by DGL. 
For all the datasets but Reddit, we solve the resulting optimization problem with the Adam algorithm (a variant of stochastic gradient descent with momentum and adaptive learning rate). For Reddit dataset we use the L-BFGS algorithm \cite{byrd2016stochastic}.
We used early stopping {(with the patience set to $100$)} and model checkpoint, monitoring the accuracy on the validation set.
We set the maximum number of epochs to $500$. All the experiments involved a shallow model composed of a single layer followed by a softmax activation function. 
For hLGC, we instantiated the function $f$ as a linear single-layer neural network: $f(\mathbf{X},\mathbf{H}^{(k-1)}) = [\mathbf{X},\mathbf{H}^{(k-1)}]\mathbf{W} + \mathbf{b}$, where $\mathbf{W} \in \mathbb{R}^{(n+m) \times m}$, $\mathbf{b} \in \mathbb{R}^{m}$. Both $\mathbf{W}$ and $\mathbf{b}$ are learned. \\
The results were obtained by performing 5 runs for each model. For our experiments, we adopted a machine equipped with: 2 x Intel(R) Xeon(R) CPU E5-2630L v3, 192GB of RAM and a Nvidia Tesla V100. For more details please check the publicly available code\footnote{\url{https://github.com/lpasa/LGC}}.

\subsubsection{Model selection}\label{sec:model_selection}
Before discussing the results of the proposed graph convolutions in the perspective of results of state-of-the-art methods, we would like to point out that for different reasons, the results reported in literature are not always comparable to the ones we report here.
For instance, there may be different versions of the same dataset (using the same name), or different train/validation/test splits on the same dataset that may significantly impact the reported results.
Another aspect to consider is the procedure adopted to select the hyper-parameters (such as learning rate, regularization, network architecture, etc.). Many papers report, for each dataset, the best performance on the test set obtained after testing many hyper-parameter configurations. This procedure favours complex methods that depend on many hyper-parameters, since they have a larger set of trials to select from compared to simpler methods. However, the predictive performances computed in this way are not unbiased estimations of the true error, thus these results are not comparable to other model selection methods.
For these reasons, we consider in this paper two experimental settings. In the first one, following many works in literature, we report the performance of the best hyper-parameter configuration for each dataset. As mentioned before, {\it these results shall be considered as an upper bound} on the predictive performance of the method. We report the results and the discussion concerning this setting in \ref{appendix:results}.
In the second experimental setting, that we discuss in the main paper, we select all the hyper-parameters of each method on the validation set. We then classify the test set with a single model.

The hyper-parameters of the model (number of hidden units, learning rate, weight decay, $k$) were selected by using a limited grid search, where the explored sets of values do change based on the considered dataset. We performed  some preliminary tests in order to select the set of values taken into account for each hyper-parameter. In Table~\ref{tab:grid_parameter}, we report the sets of hyper-parameter values used for the grid search. In order to perform a fair comparison among the proposed models and the baselines, we use the same hyper parameters grid for all the models. As evaluation measure, we used the average accuracy computed on the validation set.

\setlength\tabcolsep{1pt}
\begin{table}[t]
  \centering
  \small
   \resizebox{0.99\textwidth}{!}{
  \begin{tabular}{c|c|c|c|c|c }
      \toprule
        {\textbf{Dataset}} & {learning rate} & {weight decay} & {drop out} & {k} & {\#hidden}  \\
        \midrule
        {\textbf{Citeseer}} & $0.2$, $0.02$, $0.001$ & $10^{-2}$, $5\cdot10^{-3}$, $5\cdot10^{-4}$ & $0.0$,  $0.2$, $0.5$ & $2$,$5$,$10$,$20$,$40$,$50$,$60$ & $4$, $8$, $16$, $24$ \\
        \hline
        {\textbf{Cora}}  & $0.2$, $0.05$, $0.001$ & $5\cdot10^{-3}$,$5\cdot10^{-4}$,$5\cdot10^{-6}$, & $0.0$, $0.2$, $0.5$ & $2$,$5$,$10$,$20$,$40$,$60$,$80$& $4$, $8$, $16$, $24$\\
        \hline
        {\textbf{Pubmed}}  & $0.2$, $0.05$, $0.001$ & $5\cdot10^{-3}$,$5\cdot10^{-4}$,$5\cdot10^{-6}$, & $0.0$,  $0.2$, $0.5$ & $2$,$5$,$10$,$20$,$40$,$60$,$80$& $4$, $8$, $16$, $24$\\
        \hline
        {\textbf{Reddit}}  & $1$, $0.5$, $0.05$, $0.005$ & - &  $0.0$,$0.5$ & $2$,$4$,$6$ & $1$,$2$, $4$, $8$\\
   \bottomrule
  \end{tabular}}
  \caption{Sets of hyper-parameters values used for model selection via grid search.}
  \label{tab:grid_parameter}
\end{table}

\subsection{Experimental Results}
Table~\ref{tab:results_valid} reports the results obtained validating all the hyper-parameters on the validation set. For each method and dataset we report the average accuracy and the standard deviation over 5 runs. For sake of completeness we have also reported in \ref{appendix:results} the result obtained selecting the best hyper-parameter values on the test set.  We recall that this hyper-parameter selection procedure is biased, as discussed in the previous section.

Let us start considering the first proposed graph convolution: EGC. It shows competitive predictive performance, performing more than $2\%$ better than SGC on Citeseer and being comparable to GAT and GCN. On Cora, EGC performs comparably to SGC and GAT, and slightly worse than GCN. On the Pubmed dataset, EGC performs slightly worse than SGC, but comparably to GCN and slightly better than GAT. On Reddit, EGC is comparable to SGC, that in turn performs better than GCN. We could not compute GAT on Reddit since even using the simplest possible model (with a single attention head) the memory requirements are higher than the 16GB that are available on our GPU.

Considering LGC, 
it performs better than EGC in all the considered datasets. Moreover, 
it outperforms the competing methods in literature in all the considered datasets, including more complex nonlinear models, like GAT. Note that both GAT and GCN baselines exploit a layer with a different number of hidden units compared to SGC and the proposed EGC and LGC models, that directly compute the representation in the output space. 
Finally, the hLGC model achieves the best predictive performance in all the considered datasets.

\setlength\tabcolsep{2.5pt}
\begin{table}[t]
  \centering
  \small
  \begin{tabular}{l|cccc}
    \toprule
    {\textbf{Model} $\backslash$ \textbf{Dataset}} &  {\textbf{Citeseer}} & {\textbf{Cora}} & {\textbf{Pubmed}} & {\textbf{Reddit}}\\
     \midrule
     \textbf{SGC} & $69.2\rpm0.0$ & $80.1\rpm0.04$ & $79.8\rpm0.0$ & $94.7\rpm0.05$\\
    \arrayrulecolor{black!20}\hline
    \textbf{GAT} & $70.7\rpm0.81$ & $80.5\rpm1.02$ & $78.3\rpm0.95$ & OOM\\
    \arrayrulecolor{black!20}\hline
    \textbf{GCN} & $71.3\rpm0.48$ & $81.0\rpm0.67$ & $79.2\rpm0.63$ & $90.1\rpm0.12$\\
    
    \arrayrulecolor{black}\hline
    \textbf{EGC} & $71.3\rpm0.15$ & $80.3\rpm0.46$ & $79.4\rpm0.22$ & $94.5\rpm0.01$\\
    \arrayrulecolor{black!20}\hline
    \textbf{LGC} &  $72.2\rpm0.33$ & $82.0\rpm0.52$ & $80.6\rpm0.40$ & $95.2\rpm0.06$\\
    \arrayrulecolor{black!20}\hline
     \textbf{hLGC} & $\mathbf{72.3\rpm0.10}$ & $\mathbf{82.4\rpm0.88}$ & $\mathbf{80.8\rpm0.01}$ &  $\mathbf{95.6\rpm0.07}$\\
    \arrayrulecolor{black}\bottomrule
  \end{tabular}
  \caption{Accuracy comparison between the proposed models and three baselines (SGC, GAT and GCN). The model selection is preformed considering the results obtained on the validation set. }
  
  \label{tab:results_valid}
\end{table}

\subsection{Computational requirements}
\label{sec:times}
In Table~\ref{tab:time}, we report the average computational time required to perform a single epoch for the three considered convolutions from the literature, and our proposals.
We report, for each convolution and dataset, the computational times corresponding to the hyper-parameters that provide the best predictive results. We report the average duration (and standard deviation) of all the training epochs, in milliseconds.
We can notice that GAT and GCN are, in all the three datasets, significantly slower compared to EGC and LGC that allow to pre-compute the exponentiations of the adjacency matrix. This is not the case for hLGC due to the overhead introduced by the hyper networks used for each $i$. Notice that the reported times are an average over all the epochs, thus the pre-processing time for SGC, EGC, LGC and hLGC is included in the reported time.

\setlength\tabcolsep{2.5pt}
\begin{table}[t]
  \centering
  \small
  \begin{tabular}{l|ccc}
    \toprule
     &  {\textbf{Citeseer}} & {\textbf{Cora}} & {\textbf{Pubmed}} \\ 
     \midrule
    \textbf{SGC} & $12\rpm0.9$ & $3.6\rpm0.1$ & $3.3\rpm0.1$ \\
    \arrayrulecolor{black!20}\hline
    \textbf{(k)} & $(20)$ & $(5)$ & $(10)$ \\
    \arrayrulecolor{black}\hline
    \textbf{GAT} & $20.0\rpm.0.4$ & $20.3\rpm0.2$ & $19.3\rpm0.2$\\
    \arrayrulecolor{black!20}\hline
    \textbf{(\#hidden)} & $(8)$ & $(4)$ & $(16)$ \\
    \arrayrulecolor{black}\hline
    \textbf{GCN} & $20.4\rpm0.4$ & $20.2\rpm0.2$ & $19.4\rpm0.4$\\
    \arrayrulecolor{black!20}\hline
    \textbf{(\#hidden)} & $(24)$ & $(16)$ & $(16)$ \\
    
    \arrayrulecolor{black}\hline
    \textbf{EGC} & $8.9\rpm0.2$ & $6.8\rpm0.4$ & $4.5\rpm0.3$\\
    \arrayrulecolor{black!20}\hline
    \textbf{(k)} & $(10)$ & $(20)$ & $(20)$ \\
    \arrayrulecolor{black}\hline
    \textbf{LGC} &  $6.8\rpm0.2$ & $10.1\rpm0.4$ & $5.2\rpm0.3$\\
    \arrayrulecolor{black!20}\hline
    \textbf{(k)} & $(10)$ & $(20)$ & $(20)$ \\
    \arrayrulecolor{black}\hline
    \textbf{hLGC} & $52.4\rpm4.3$ & $50.8\rpm0.3$ & $35.1\rpm0.8$\\
    \arrayrulecolor{black!20}\hline
    \textbf{(k)} & $(10)$ & $(80)$ & $(40)$ \\
    \arrayrulecolor{black}\bottomrule
  \end{tabular}
  \caption{Time comparison among the model proposed in this paper and three baseline (SGC, GAT, and GCN).For each convolution and dataset, we report the average duration (and standard deviation) of all the training epochs of the best performing model.} The time measurements are reported in milliseconds.
  
  \label{tab:time}
\end{table}
\subsection{Discussion}
While the improvement of hLGC compared to the second best performing method (LGC) on each dataset seems marginal, it is worth to notice that hLGC consistently performs better than other methods.
While all three proposed methods perform consistently better than SGC, hLGC is the method showing the best predictive performance, while LGC exhibits the best trade-off among predictive performance and required computational time.

\subsection{Comparison among simple convolutions}
\begin{figure}
  \begin{subfigure}{}
  \includegraphics[width=.50\linewidth]{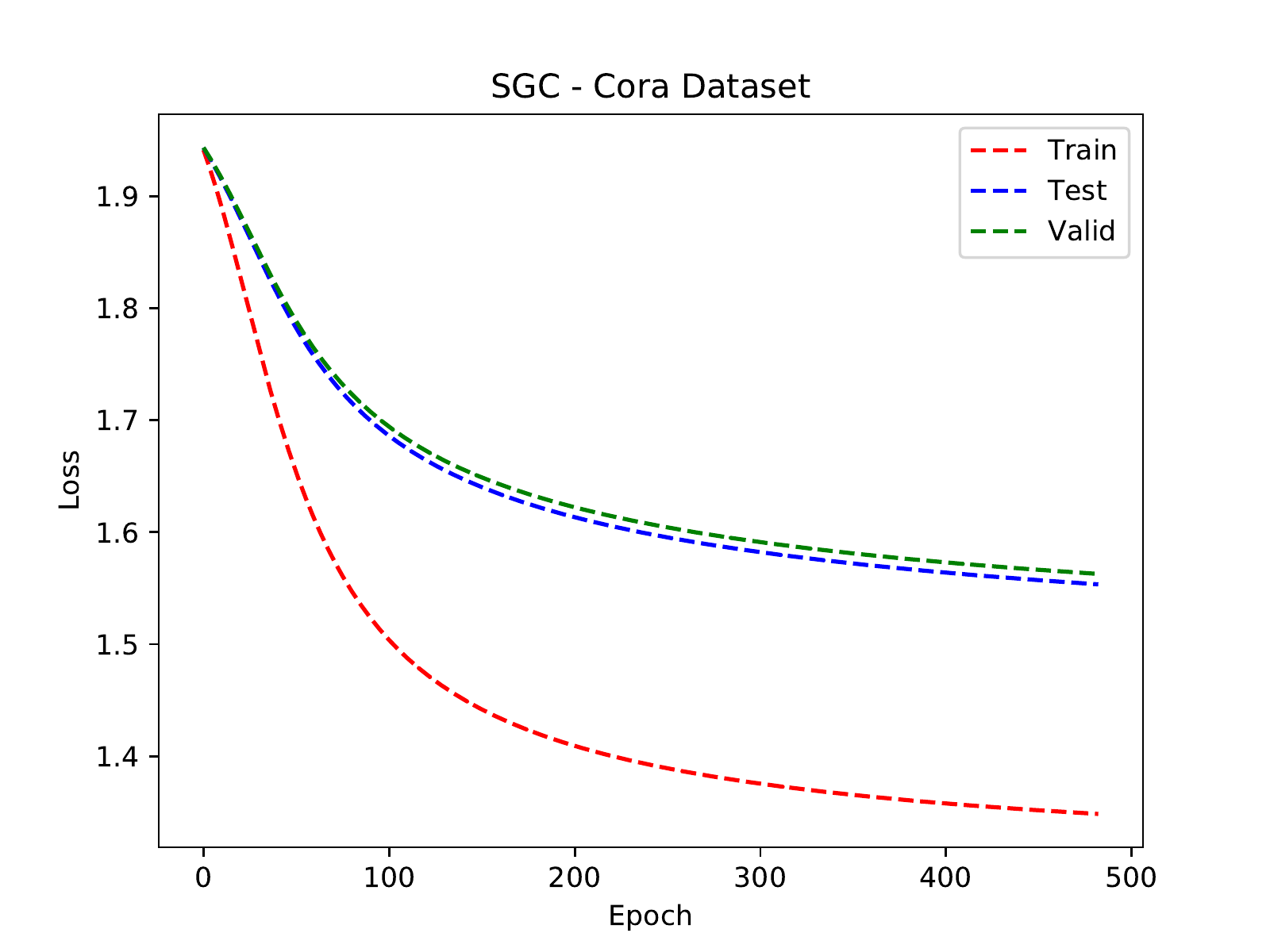}\hfill
  \includegraphics[width=.50\linewidth]{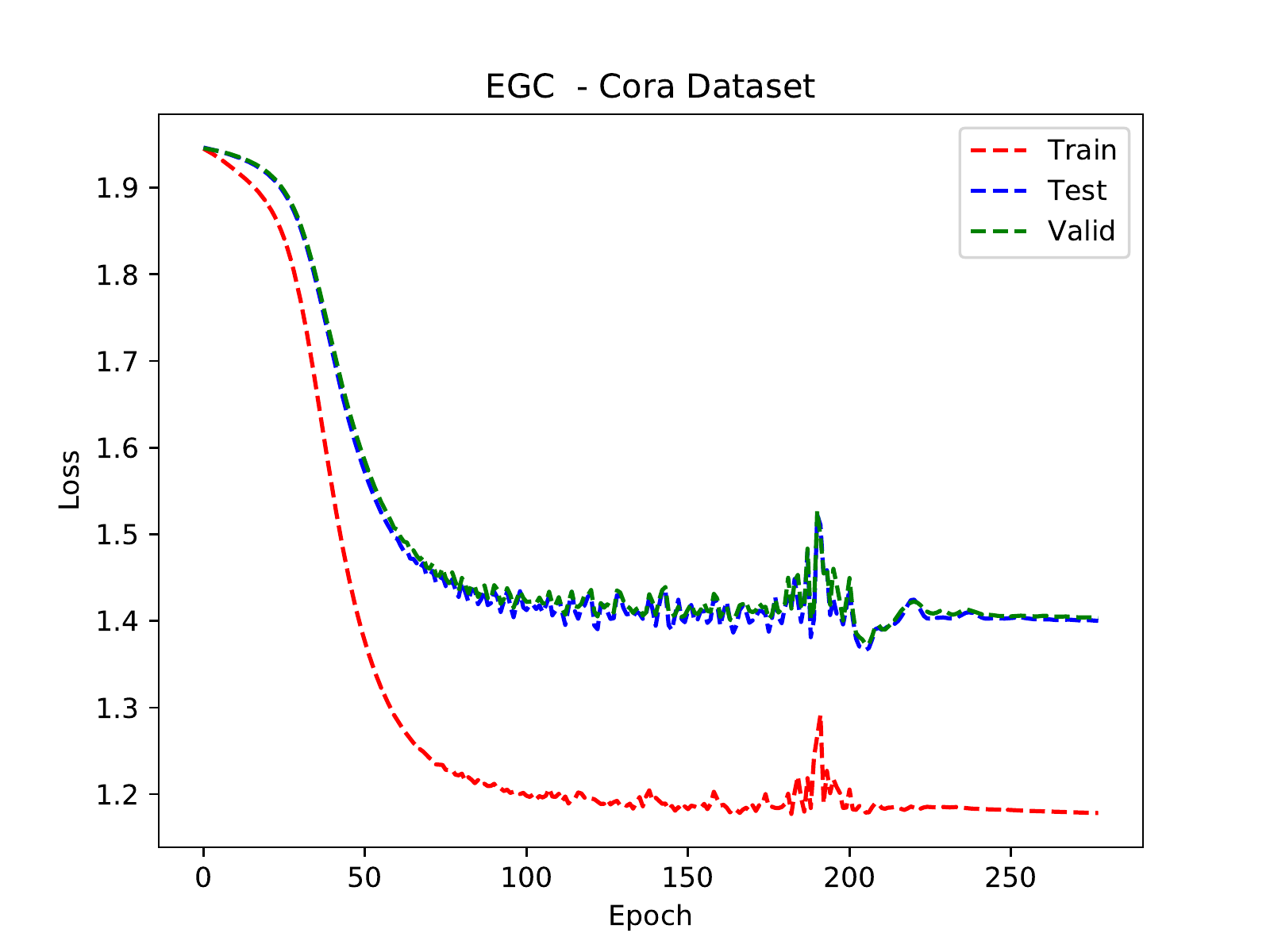}\hfill\\
  \includegraphics[width=.50\linewidth]{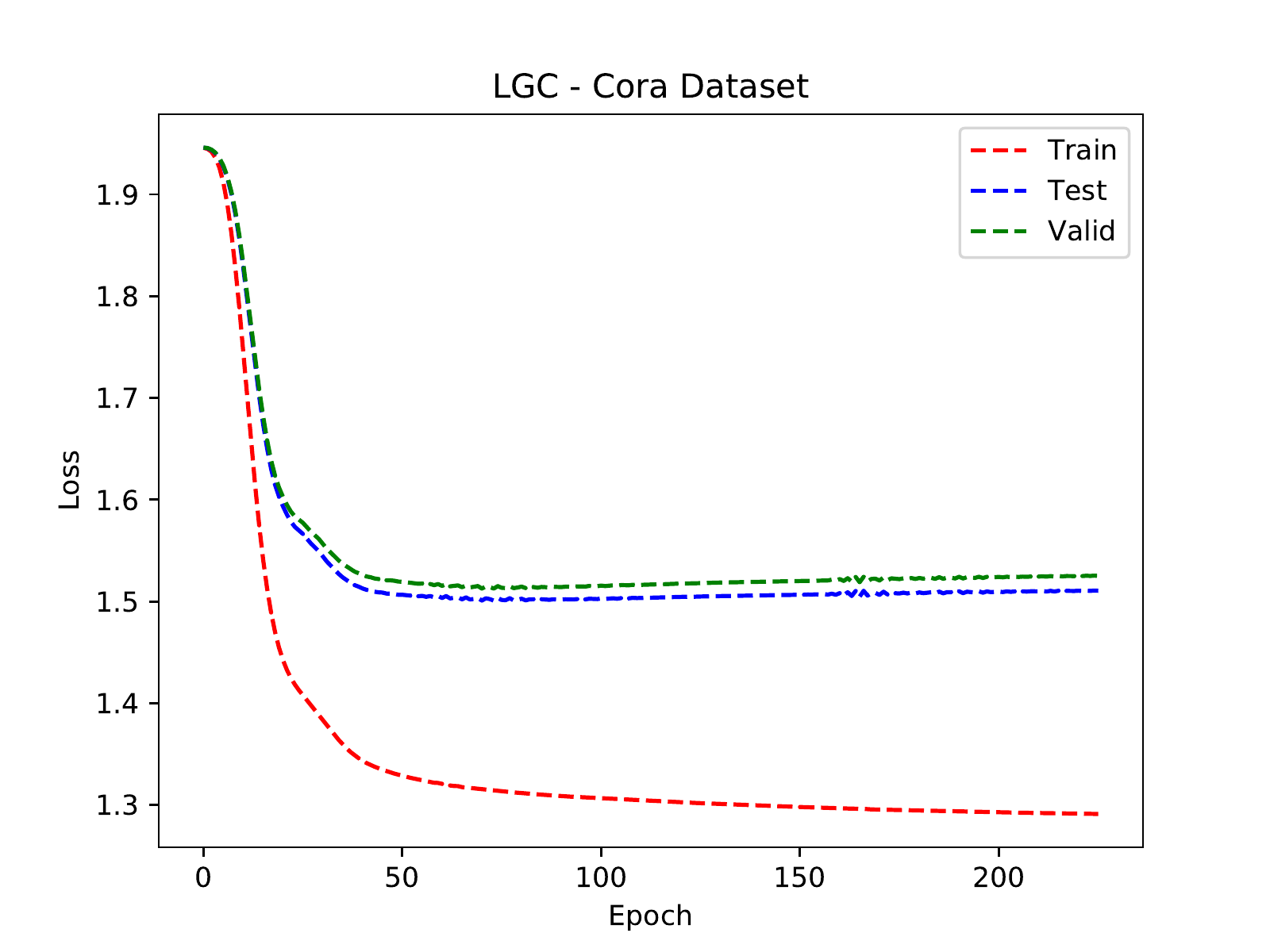}\hfill    
  \includegraphics[width=.50\linewidth]{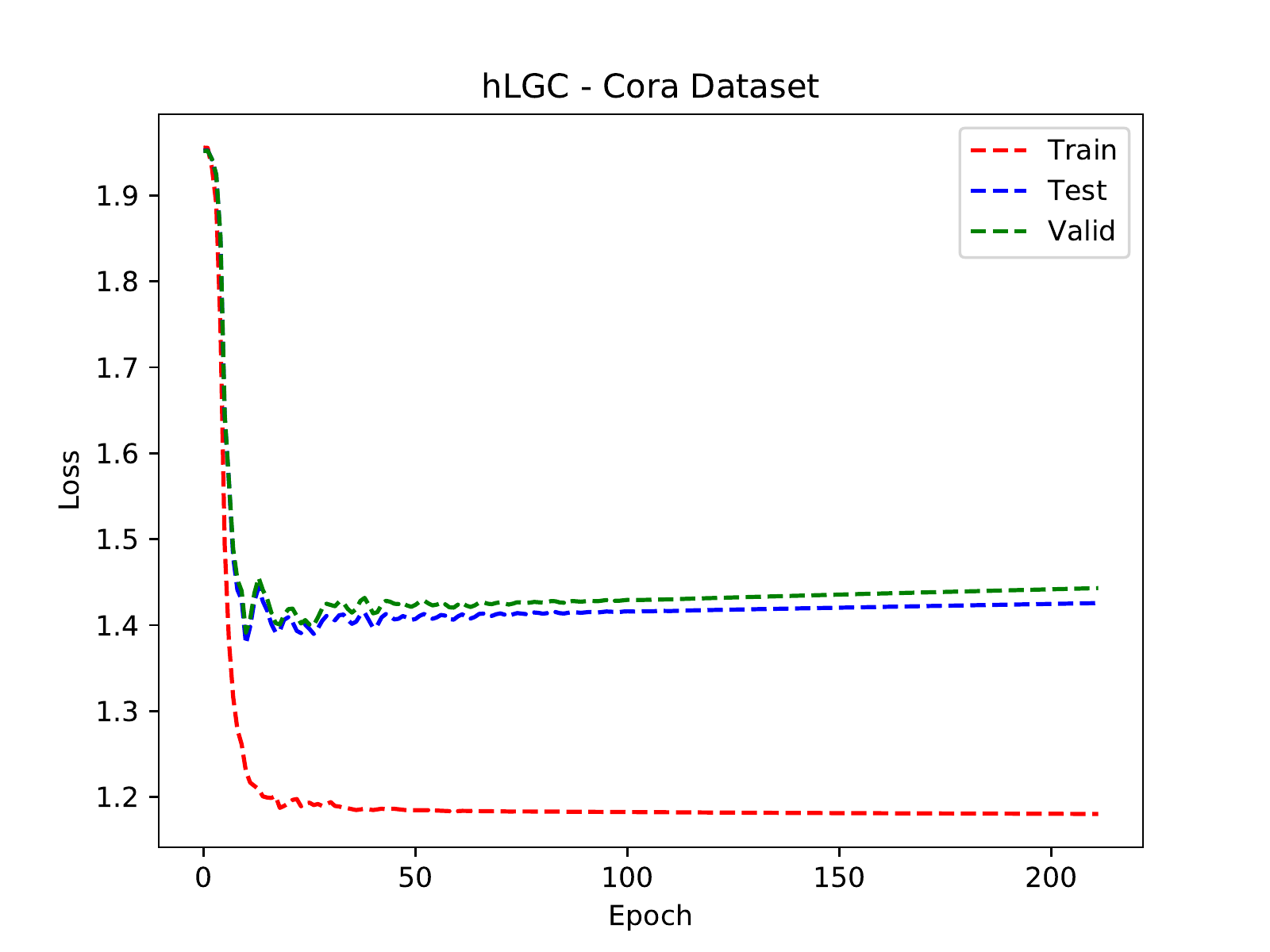}\hfill
  \end{subfigure}\par\medskip
    \caption{Loss curves computed with the progress of training epochs on Cora training, test, and validation sets for SGC, EGC, SGC and hLGC.}
    \label{fig:curve}
\end{figure}

In Figure~\ref{fig:curve}, we report loss curves during training using SGC and the three convolutions proposed in this work (EGC, LGC and hLGC) on the Cora dataset. On the other datasets, similar considerations can be drawn. The plots report the loss computed on the training, validation, and test sets. The curves refer to the hyper-parameters that yield the best results for each method (whose results are reported in Table~\ref{tab:results_no_valid}).  It is interesting to notice how the number of epochs that the model requires to converge is related to the expressiveness of the considered convolution. In fact, for SGC, EGC, LGC and hLGC (that are increasingly expressive) we can see that the slope of the curves becomes steeper as a more expressive model is used. 

\begin{figure}
\includegraphics[scale=0.4]{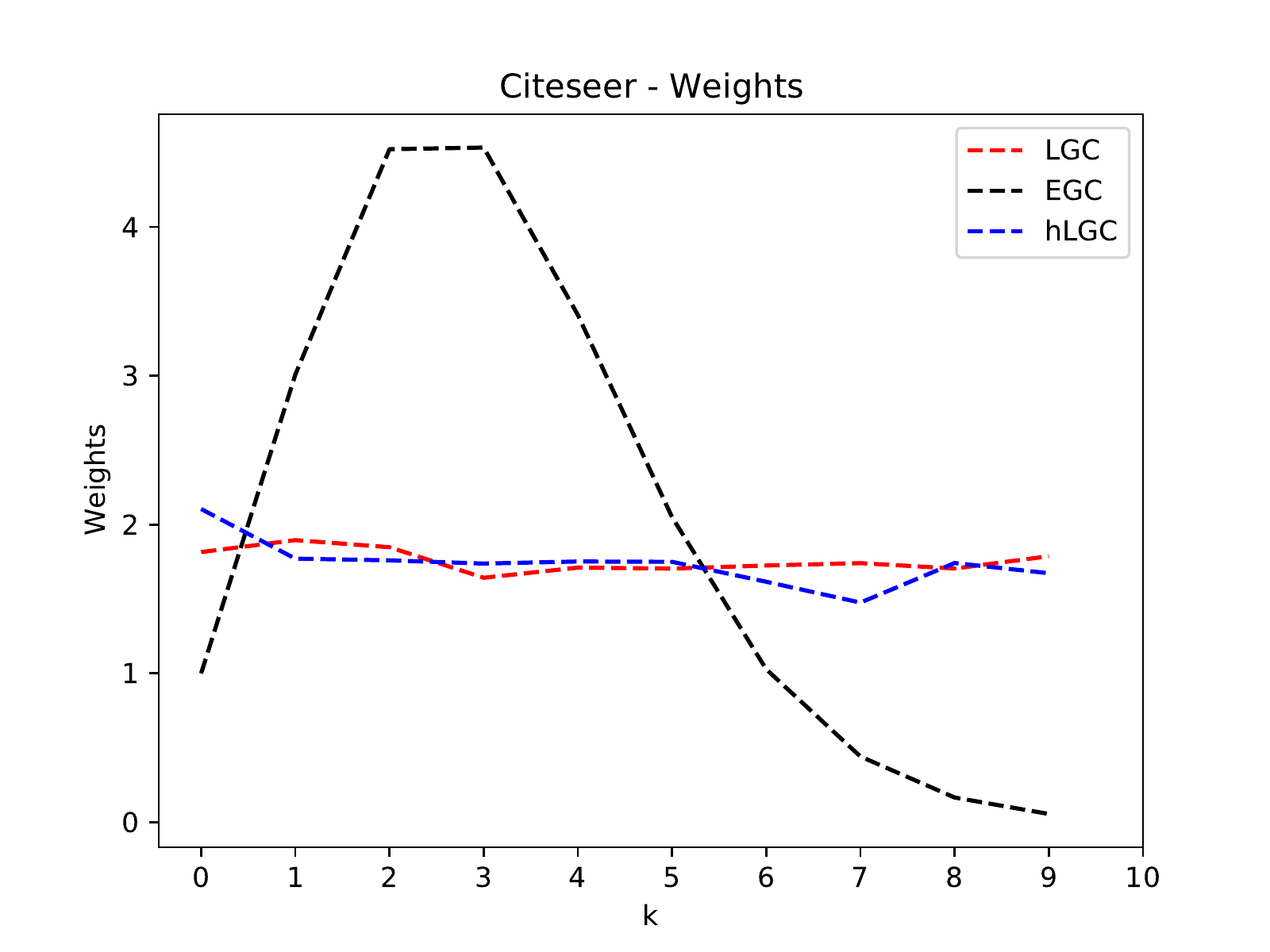}
\includegraphics[scale=0.4]{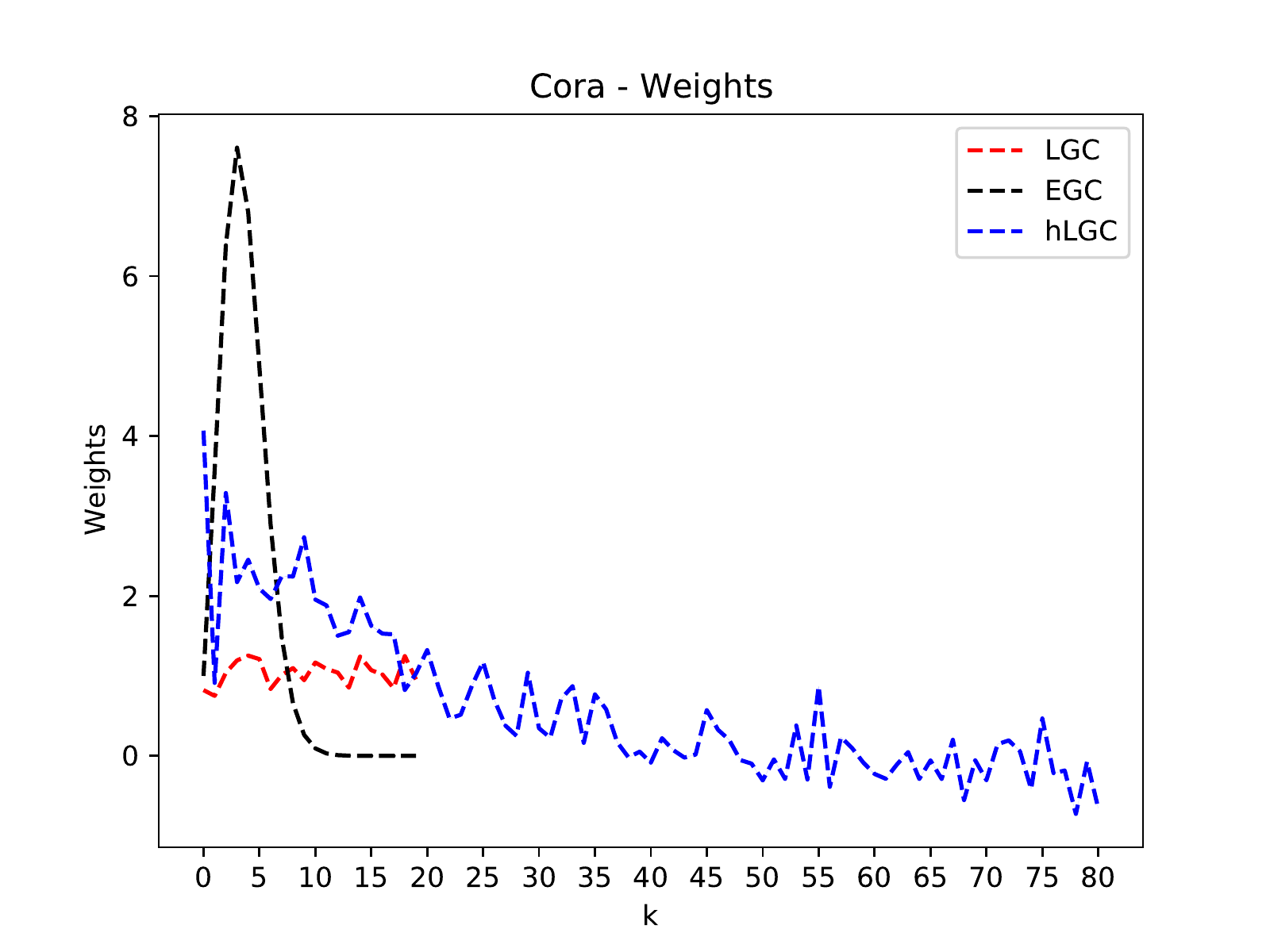}\\
\centering
\includegraphics[scale=0.4]{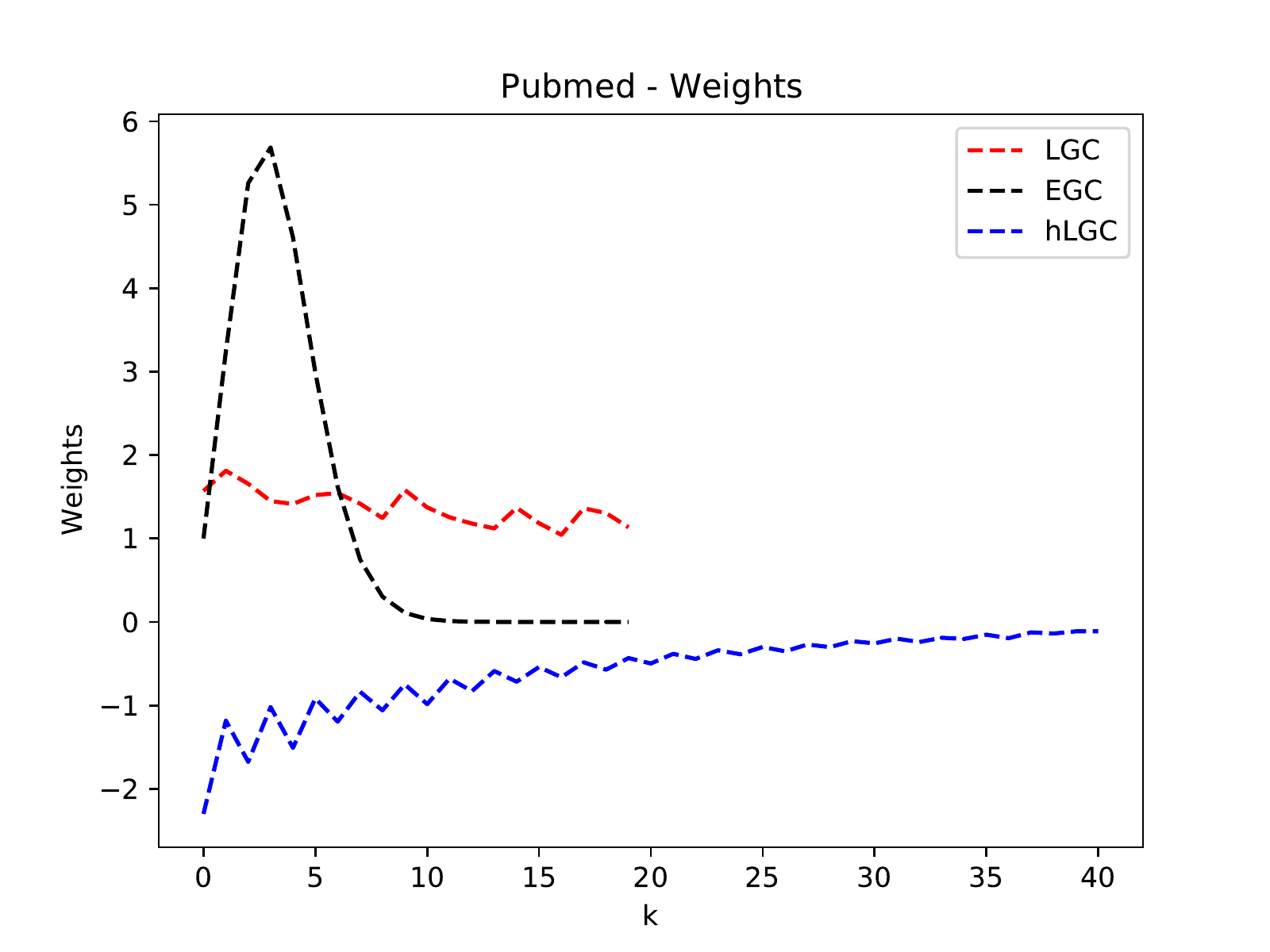}
\caption{Values of the $\alpha_i$ parameters of the LGC, values of $\frac{\beta^i}{i!}$ on EGC and, (averaged) values of $f_i(\mathbf{L}^i\mathbf{X})$ on hLGC (variance is reported as well, but it is too small to be visualized), for different values of $k$. The considered models are trained on the Citeseer, Cora and Pubmed datasets and their hyper-parameters are selected on the respective validation sets.}
 \label{fig:alpha_curve}
\end{figure}

An important role in the three proposed convolutions is played by the multiplicative 
coefficients
applied to each term of the summation, which  significantly influence the optimization phase and the final results. For this reason, we decided to study the values of these multiplicative elements for all new models on the Citeseer, Cora, and Pubmed datasets. In Figure~\ref{fig:alpha_curve}, we have reported the coefficient values for the three  models selected in validation. For what concerns EGC, a single learned parameter $\beta$ determines the weight of each $i \in \{0, \dots, k\}$ term of the summation (see eq.~\eqref{eq:EGC}), 
computed as $\frac{\beta^i}{i!}$. These values are represented by the black line in Figure~\ref{fig:alpha_curve}. The LGC convolution defines, instead, a different multiplicative coefficient $\alpha_i$ for each  $i \in \{0, \dots, k\}$ (red line in Figure~\ref{fig:alpha_curve}). All the $\alpha_i$ (similarly to $\beta$) are adjusted during the optimization.
Finally, using the blue line, we report the average of the output of the hyper networks $f_i(\mathbf{L}^i\mathbf{X})$ for each value of $i \in \{0, \dots, k\}$ for the hLGC model. Variance is also reported, however it is so small that it is not possible to discriminate it in the plot.
We can notice that the coefficients learned by LGC  tend to be closer to each other compared to EGC, while the (average) coefficients generated in the hLGC show a much larger range of variation and diversification with respect to the other two models. 
From these plots it is evident that EGC is much more constrained w.r.t. LGC, being forced to concentrate significant values on few nearby terms. In addition to that, the hLGC model selected in validation exploits a much larger value of $k$, thus showing a better ability to extract significant information from large receptive fields on the graph. Finally, the very small variance observed for the output of the hyper networks seems to be an indication that the selected model does not overfit the training data.

\section{Conclusion and Future Directions}
In this paper, we followed the opposite direction compared to many works in literature on the definition of graph convolution operators. Instead of increasing the complexity of existing options, we started from graph spectral filtering theory, and defined three increasingly expressive graph convolutions. For two of these models, i.e. EGC and LGC, we also provided Rademacher generalization bounds that, due to the simplicity of the proposed models, can be directly applied.
We showed that our proposals achieve state-of-the-art predictive performance while being more efficient (EGC and LGC) to compute than most alternatives in literature.

In the future, we plan to expand the study of the Rademacher complexity bounds on other graph convolutions and analyze if the bounds are tight enough to allow for a comparison of the expressiveness of different graph operators.
Moreover, we would like to study the effects of stacking multiple graph convolution layers in an architecture that include non-linearities. Finally, we plan to test the proposed convolutions in the setting of graph classification (instead of node classification, considered in this paper).

\section{Acknowledgements}
The authors acknowledge the HPC resources of the Department of Mathematics, University of Padua, made available for conducting the research reported in this paper.
This work was partly funded by the SID project (BIRD 2020) ``Deep Learning for Graph Memory Networks" CUP C99C20001480005.

\bibliographystyle{elsarticle-num}
\bibliography{references}
\newpage
\appendix

\section{Results - hyper-parameter selection on test set}\label{appendix:results}
The results reported in Table~\ref{tab:results_no_valid} were obtained selecting the best hyper-parameter values on the test set. For each hyper-parameter configuration, the model with highest validation accuracy was selected. We recall that this hyper-parameter selection procedure is biased, as discussed in the section \ref{sec:model_selection}. We can notice that the performance obtained in the validated setting in Table~\ref{tab:results_valid} are in general lower compared to the ones in Table~\ref{tab:results_no_valid}. Complex methods such as GAT tend to show the higher decrease in accuracy.
\setlength\tabcolsep{2.5pt}
\begin{table}[t]
  \centering
  \small
  \begin{tabular}{l|cccc}
    \toprule
    {\textbf{Model} $\backslash$ \textbf{Dataset}} &  {\textbf{Citeseer}} & {\textbf{Cora}} & {\textbf{Pubmed}} & {\textbf{Reddit}}\\
     \midrule
     \textbf{SGC} & $70.70\rpm0.0$ & $81.1\rpm0.0$& $79.8\rpm0.0$ & $94.7\rpm0.05$\\
     \arrayrulecolor{black!20}\hline
     \textbf{GAT} & $71.1\rpm0.37$ & $82.2\rpm0.81$ & $78.6\rpm0.24$ & OOM\\
     \arrayrulecolor{black!20}\hline
     \textbf{GCN} &$71.6\rpm0.41$ & $81.3\rpm0.43$ & $79.3\rpm0.62$ & $90.1\rpm0.29$\\
   
    \arrayrulecolor{black}\hline
    \textbf{EGC} & $71.8\rpm0.10$ & $81.2\rpm1.0$ & $79.6\rpm0.13$ & $94.5\rpm0.01$\\
    \arrayrulecolor{black!20}\hline
    \textbf{LGC} &  $72.2\rpm0.16$ & $82.6\rpm0.07$ & $81.0\rpm0.10$ & $95.3\rpm0.03$\\
    \arrayrulecolor{black!20}\hline  
    \textbf{hLGC} & $\mathbf{73.2\rpm0.02}$ & $\mathbf{83.0\rpm0.01}$ & $\mathbf{81.2\rpm0.03}$ &  $\mathbf{95.6\rpm0.03}$\\
    \arrayrulecolor{black}\bottomrule
  \end{tabular}
  \caption{Accuracy comparison between the proposed models and three baselines (SGC, GAT and GCN). The model selection is preformed considering the results obtained on the test set. }
  
  \label{tab:results_no_valid}
\end{table}
\end{document}